\documentclass[sigconf]{acmart}

\usepackage{hyperref}
\usepackage{url}
\usepackage{booktabs}
\usepackage{multirow}
\usepackage{float}
\usepackage[ruled,vlined,linesnumbered]{algorithm2e}
\usepackage{graphicx}
\usepackage{subfigure}
\usepackage{balance}

\AtBeginDocument{%
  }

\setcopyright{acmlicensed}
\copyrightyear{2018}
\acmYear{2018}
\acmDOI{XXXXXXX.XXXXXXX}

\acmConference[Conference acronym 'XX]{Make sure to enter the correct
  conference title from your rights confirmation email}{June 03--05,
  2018}{Woodstock, NY}

\acmISBN{978-1-4503-XXXX-X/2018/06}

\begin{document}

\title{Multi-modal Rail Crossing Safety Analysis}

\author{Paimon Goulart}
\authornote{Three authors contributed equally to this research.}
\affiliation{%
  \institution{University of California, Riverside}
  \city{Riverside}
  \state{California}
  \country{USA}
}

\author{Chansong Lim}
\authornotemark[1]
\affiliation{%
  \institution{University of California, Riverside}
  \city{Riverside}
  \state{California}
  \country{USA}
}

\author{Nícolas Roque dos Santos}
\authornotemark[1]
\affiliation{%
  \institution{University of California, Riverside}
  \city{Riverside}
  \state{California}
  \country{USA}
}

\author{Yue Dong}
\affiliation{%
  \institution{University of California, Riverside}
  \city{Riverside}
  \state{California}
  \country{USA}
}

\author{Sheldon Peterson}
\affiliation{%
  \institution{Riverside County Transportation Commission}
  \city{Riverside}
  \state{California}
  \country{USA}
}

\author{Jia Chen}
\affiliation{%
  \institution{University of California, Riverside}
  \city{Riverside}
  \state{California}
  \country{USA}
}

\author{Evangelos E. Papalexakis}
\affiliation{%
  \institution{University of California, Riverside}
  \city{Riverside}
  \state{California}
  \country{USA}
}

\renewcommand{\shortauthors}{Trovato et al.}

\begin{abstract}
Given one or more images of a railway crossing, can we leverage visual cues that allow us to robustly estimate how safe it is? Can we improve our ability to do so by introducing structured data (such as official accident reports) about the accident history of that crossing into our models? In this work, we explore how to best answer those questions towards building an AI system that can ingest multi-modal data for railway crossings and provide safety assessment and scores that align with expert opinion and with safety scoring used by the Federal Railroad Administration (FRA). To that end, we propose a proof-of-concept pipeline that delivers on that goal, while at the same time exploring and tackling a number of critical research challenges that pertain to different parts of the pipeline, from data preparation to different learning paradigms that can allow us to realize such a system. Indicatively, our proposed system identifies HIGH-RISK and LOW-RISK crossings with a macro F1 score of 0.757 and estimates FRA-based safety scores with an RMSE of 0.078 and correlation of 0.492 using a routed fine-tuned compact VLM pipeline, while producing qualitative results that align with domain-expert assessment.
\end{abstract}



\received{20 February 2007}
\received[revised]{12 March 2009}
\received[accepted]{5 June 2009}

\maketitle

\section{Introduction}
\label{sec:intro}
Highway–rail grade crossings are a critical safety concern in United States transportation systems. Each year, more than 2,000 collisions and over 200 fatalities occur at these crossings, making them the second leading cause of rail-related deaths nationally~\cite{fra_crossing_safety}. Therefore, improving crossing safety requires reliable ways to assess risk and prioritize interventions. However, this is challenging because crossing risk depends on multiple interacting factors, including railway traffic, roadway characteristics, crossing infrastructure, weather conditions, visibility, and the behavior of highway users. In practice, assessing these factors often requires expert knowledge and careful inspection of crossing conditions. Given the large number of crossings and the limited expert attention available at any point in time, manually assessing crossing safety is difficult to scale, motivating AI-assisted tools for systematic crossing safety assessment.

Existing railway crossing risk assessment methods primarily rely on statistical models. For example, the Accident Prediction and Severity (APS)~\cite{APS_score, new_APS_score} model provides a standardized framework for estimating crossing risk, and its outputs are made available through the Federal Railroad Administration's (FRA) Grade Crossing Accident Prediction System (GXAPS)~\footnote{GXAPS is available at \url{https://railroads.dot.gov/railroad-safety/divisions/crossing-safety-and-trespass-prevention/grade-crossing-accident}}. This tool supports the identification of high-risk crossings and help guide the allocation of limited safety resources. However, because APS depends on predefined crossing variables (e.g., number of daily trains and average annual daily traffic), they may not fully capture visual and context-dependent factors observed by highway users approaching a crossing. At the same time, prior work has shown that inaccurate or incomplete crossing inventory data can significantly affect crash and severity predictions, leading to differences in estimated risk and crossing rankings~\cite{farooq2023effects}.

These challenges motivate the use of visual information as a complementary source of evidence for crossing safety assessment. This is especially relevant for highway–rail grade crossings, where safety depends not only on crossing attributes recorded in databases, but also on what highway users can perceive as they approach the tracks. Street-level imagery can represent this perspective and reveal safety-relevant cues that are difficult to encode only through structured fields and predefined variables, such as the visibility of warning devices, sightlines, visual obstructions, road geometry, weather effects, and surrounding conditions. Thus, this visual perspective can support risk assessment by capturing what is observable to a highway user as they approach a crossing.

Reasoning over these visual observations requires models that can understand complex scenes and related visual cues to safety-relevant concepts. Recently, Vision-Language Models (VLMs) have been explored in a diverse set of safety-related assessment tasks, including railway intrusion perception~\cite{cograil}, safety-oriented reasoning in train cab-view scenes~\cite{railqva}, road safety assessment~\cite{vroast}, and workplace hazard analysis~\cite{mining_safety, construction_hazards}. These studies suggest that VLMs can reason over complex visual scenes and produce descriptions of safety-critical conditions. However, these efforts have not yet addressed highway–rail grade crossing safety from the highway user’s perspective, where risk depends on how drivers perceive crossing infrastructure and the surrounding environment.

To address this gap, we investigate whether VLMs can support scalable highway–rail grade crossing safety analysis by connecting street-level visual evidence, historical incident records, and official risk scoring. Specifically, we introduce a multimodal pipeline that combines crossing image sequences with FRA Form 57 records and evaluates VLMs in two complementary tasks: \textit{Risk Scoring} and \textit{Visual Risk Inspection}.

In the risk scoring task, we evaluate if VLMs can estimate APS crossing risk scores from multimodal crossing information. We consider image-only prompting, prompting with both images and Form 57 records, and model fine-tuning. We study this problem both as a binary risk classification task, where the model identifies whether a crossing is low- or high-risk, and as a continuous score prediction task, where the model estimates the annual average predicted accident score. To better handle the skewed distribution of risk scores, we further evaluate a routed prediction strategy that separates crossings into risk groups before estimating their scores.

In the visual risk inspection task, we examine whether VLMs can identify safety-relevant visual factors from the highway user's perspective. Rather than asking the model to describe risks in an unconstrained way, we ground the analysis in accident theory and use predefined categories of threats, safety infrastructure, and escalation factors. We also evaluate the model under augmented weather and illumination conditions, such as rain, strong sunlight, and nighttime, to assess how visual conditions may affect the visibility of warning devices and other crossing features.

Our results show that VLMs are promising for railway crossing safety analysis, with fine-tuning improving their ability to estimate risk from multimodal crossing data. Beyond score prediction, the models also show potential for identifying visual safety factors from the highway user's perspective, highlighting their capacity to capture visual evidence related to crossing safety.

Our main contributions are as follows:
\begin{itemize}
    \item We propose a multimodal pipeline for highway–rail grade crossing safety assessment that combines street-level crossing imagery with FRA Form 57 incident records.
    \item We study VLM-based risk score estimation, where the model predicts FRA-style safety scores from multimodal crossing information.
    \item We provide quantitative and qualitative analyses of how VLMs can support railway crossing safety assessment beyond structure scoring variables alone.   
\end{itemize}

\begin{figure*}[h]
    \centering
    \includegraphics[width=1\linewidth]{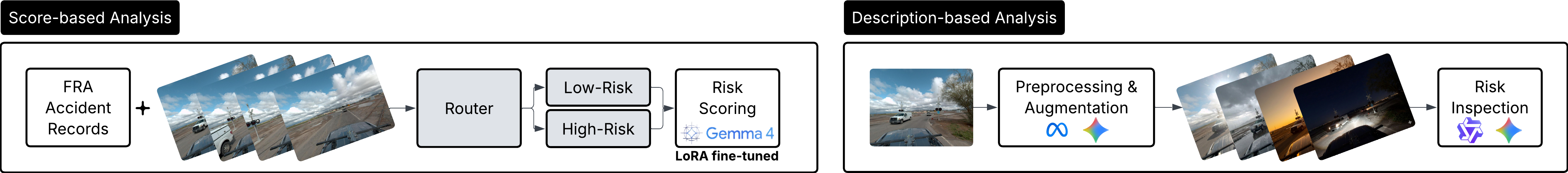}
    \caption{Overview of our proposed pipeline. In the score-based analysis, crossing imagery and FRA accident records are used to estimate risk scores, which are evaluated against FRA's official scoring system. In the description-based analysis, images are preprocessed augmented to simulate different weather and illumination conditions, such as rain and nighttime, allowing the VLM to inspect visual risk factors such as occluded signals, poor signal visibility, or objects blocking the driver’s view.}
    \label{fig:pipeline}
\end{figure*}

\section{Related Work}
\label{sec:related_work}
\subsection{Railway Risk Scoring}
For the past few decades, the United States Department of Transportation (DOT) has been using Accident Prediction and Severity (APS) model~\cite{APS_score} to identify high-risk crossings, assess its risk level, and predict potential accidents based on formula developed using nonlinear regression on crossing characteristics such as the total number of through trains per day. However, due to its lack of consistency and low prediction rate, DOT has been periodically adjusting the coefficients of the model and exploring additional methods to support it, including High-Speed Rail (HSR) Accident Severity Model~\cite{HSR_score} and supplementary safety measure impacts from Train Horn Rule (49 CFR 222), which requires locomotive engineers to stound the train horn before reaching a public highway-rail grade crossing~\cite{train_horn_rule1,train_horn_rule2}.

Despite these efforts, as rail and highway dynamics, such as crossing environments and vehicle trends, significantly change over time, DOT had to redesign the previous APS model into a new formulation~\cite{new_APS_score}. The prediction model of the newly developed method adopts Zero-Inflated Negative Binomial (ZINB) regression to properly deal with the crossings where no accidents have ever happened and Empirical Bayes (EB) to adjust regression results that are biased to the mean. They also refined the severity model by calibrating it with the probabilities of 3 categories: fatal, injury, and property damage only.

The new model was validated by comparing it against the previous APS model, demonstrating 30\% higher accuracy in predicting accidents at high-risk crossings. It can be used to estimate or simulate differences in risk following prospective infrastructure improvements to crossings or other environmental changes. The new severity model also revealed several factors that significantly determine the fatality of accidents, such as whether the crossing is located in a rural or urban area, maximum timetable speed, and the number of daily trains.
\subsection{VLMs for Risk Assessment and Safety}
Railway-specific applications of VLMs remain relatively limited, but recent work has begun to explore the use of VLMs for railway safety, particularly in scenarios that require reasoning beyond object detection or classification. For example, CogRail~\cite{cograil} introduces a benchmark for cognitive railway intrusion perception, where VLMs are evaluated on tasks such as position perception, movement prediction, and threat analysis for potential intrusion targets in railways. Similarly, RailVQA~\cite{railqva} explores visual question answering (VQA) in train cab-view scenarios, with an emphasis on safety-related questions that require the model to perform reasoning about railway environments. 

Beyond railway safety, VLMs have also been applied to road safety assessment and other safety-critical domains. V-Roast~\cite{vroast} formulates road safety assessment as a VQA task, using VLMs to inspect street-level imagery and infer road safety attributes without requiring task-specific training data. ScVLM~\cite{scvlm} enhances VLMs for understanding safety-critical events such as crashes and near-crashes by combining supervised learning, contrastive learning, and language-based event description. SeeUnsafe~\cite{seeunsafe} further studies video-based traffic accident analysis, using multimodal large language model agents to support accident-aware video classification, structured accident descriptions, and visual grounding. More broadly, related work has applied VLMs for construction safety, mining operations safety, and other hazard identification tasks~\cite{mining_safety, construction_hazards}. 

Taken together, these studies demonstrate the potential of VLMs for identifying risks and producing interpretable descriptions of safety-critical events. However, existing railway-oriented work primarily focuses on train operation or intrusion-perception settings, while broader safety applications focus on road or workplace hazard inspection. In contrast, our work studies highway-rail grade crossings from the highway user's perspective, where safety depends not only on railway infrastructure, but also on the weather conditions, layout, and environmental context observed by drivers approaching the crossing.

\section{Data Sources}
\label{sec:preliminaries}
Our pipeline relies on two complementary sources of information: publicly available safety-related railway crossing events and street-level images. First, we describe the Highway–Rail Grade Crossing Accident/Incident Report (Form 57), which provides historical records of safety-related events at highway–rail grade crossings reported to the FRA. Then, we detail the railway crossing image dataset and the filtering procedure used to construct the final set of crossings considered in our multimodal system.\\

\noindent \textbf{Highway–Rail Grade Crossing Incident Data (Form 57). }
%
Form 57\footnote{More information about Form 57 and the existing incident reports are available at \url{https://data.transportation.gov/stories/s/Highway-Rail-Crossing-Incidents-Landing-Page/9hnw-nt86/}} provides official accident and incident records reported to the FRA for events occurring at railway crossings. Each record corresponds to a reported safety-critical event involving railroad on-track equipment and a highway user, such as a car, pedestrian, or other road user. To characterize these events, Form 57 contains detailed information about the crossing location, date and time, weather conditions, highway users involved, railroad equipment involved, warning devices present at the crossing, and reported casualties.\\

\noindent \textbf{Railway Crossing Images. }
We use a dataset of street-level images of railway crossings throughout California, collected from the open-source platform Mapillary~\cite{mapillary}, as the visual input to our system. Since Mapillary is crowd-sourced, the images vary in format and acquisition conditions, including camera orientation, focal length, viewpoint, and geolocation accuracy. To obtain consistent image sequences with full views of each crossing, we retain only 3D panorama images captured with spherical cameras and reorient them toward the crossing view. The raw dataset contains 3285 images from 289 crossings at a resolution of 960x720. Importantly, each crossing is represented by multiple images, allowing us to capture the crossing from different distances and stages along the approach to the tracks.


Since our goal is to evaluate whether incident history can improve VLM-based safety assessment, we align the image dataset with Form 57 records at the crossing level. Crossings without corresponding Form 57 information are removed, resulting in a final dataset containing only crossings for which both visual observations and historical incident records are available. After filtering, the resulting dataset contains 1634 images from 149 crossings, associated with 406 forms recorded between 1975 and 2023.


\section{Proposed Pipeline}
\label{sec:pipeline}

In this section, we present our pipeline for multimodal railway crossing risk assessment, illustrated in Figure~\ref{fig:pipeline}. Given one or more street-level images of a railway crossing, the pipeline leverages VLMs to perform two complementary tasks: estimating a risk score that aligns with the APS score and identifying risk factors from the highway user’s perspective that are not reflected in the APS scoring function.

\subsection{Score-based Analysis}

The goal of score-based analysis is to identify one or more models that can autonomously determine whether a railway crossing is high risk or low risk, and estimate that crossing's safety score from large, complex multimodal data. For each crossing, we aim to provide the model with a sequence of street-level images captured within 20 meters of a crossing. These images are ordered in the direction of travel towards the center of the crossing, giving the model a video-like view of what a person sees while driving through a crossing. From this, the model is able to observe visible safety features such as warning devices, signage, road layout, visibility, and surrounding infrastructure.

However, a concern of ours is that image sequences alone may not be sufficient for this task. Several variables that are needed in order to compute the FRA-style accident prediction are not directly inferable from the image sequences alone, such as historical accident information. To incorporate this missing context without giving the model too much information, we also provide the model with Form 57 records which contain details of a specific historical accident that happened at a given crossing.

Since this task involves high resolution image sequences, specifically formatted textual records, and complex numerical risk calculations, we need a model that can handle multimodal inputs while still remaining practical enough to deploy at a low cost, and is easily reproducible. For these reasons we opt to use Google's Gemma 4 model as our base VLM~\cite{gemma4modelcard}. Gemma 4 is especially appealing as it fits all our criteria: it is openly available, supports multimodal inputs, and provides strong reasoning capabilities while remaining practical to deploy. Specifically, we use \texttt{Gemma $4$ E$4$B} in our experiments.

At the same time, prompting an off-the-shelf VLM is unlikely to be sufficient for this task because the model must combine visual evidence, accident-report context, and numerical reasoning while producing a highly constrained output format. Therefore, we fine-tune the model on the railway image sequences and accident reports (Form 57) using Low-Rank Adaptation (LoRA)~\cite{hu2021loralowrankadaptationlarge}. Prior work has shown that this parameter-efficient adaptation strategy achieves competitive performance with substantially lower computational cost than full fine-tuning. This makes LoRA very well suited for our problem where we want to adapt a compact multimodal model, without the need to spend much time or resources to retrain the model. 

In our score-based pipeline, the fine-tuned model receives the crossing image sequence and the associated Form 57 information as input, and is  trained to output the annual average predicted accidents for that crossing in a specific format. The ground-truth annual average predicted accident score for each crossing is obtained from the FRA Grade Crossing Accident Prediction System (GXAPS), which provides official APS-based risk estimates for highway–rail grade crossings. We compare this fine-tuned model against prompting-only baselines, including models that receive images alone, models that receive both images with reports, and an "oracle" model which receives images with reports as well as additional variables (e.g., annual average daily traffic (AADT), total trains, and train speed) that are not directly inferable from either the images or reports. From this, we are able to incrementally see how much each source of information helps the model in making it's final prediction. It also allows us to see how much fine-tuning is able to improve the performance of the model for our task, and even allows us to observe if the fine-tuning process is able to enable the model to learn useful proxy relationships from the available multimodal evidence, even when some scoring variables are not explicitly observed. 

In addition to predicting the exact accident score, we also formulate the task as a binary risk classification problem. This is motivated by the fact that the annual accident scores are very biased. Most crossings have a very low accident score, whereas high-risk crossings are much rarer. Due to this long-tailed distribution, a model trained only to predict continuous scores may learn a trivial bias towards low values leading to a reasonable average error, but poor ability to predict high-risk crossings. To address this. we trained a binary risk classifier that predicts whether each crossing is LOW-RISK or HIGH-RISK, where HIGH-RISK corresponds to crossings with annual average predicted accidents above a threshold of $0.09$. This allows us to test whether the model can identify crossings that are a high-risk, since predicting the exact continuous score could be difficult.

Finally, we introduce a routed regression method that combines these two ideas. Instead of using a single model to predict accident scores for all crossing, we first use the binary risk classifier as our router. The router assigns each crossing to either the LOW-RISK or HIGH-RISK group, which is then sent to a crossing risk specific regression model trained for that setting. Specifically our low-risk regressor specializes in crossings with lower accident scores and the high-risk regressor specializes in higher accident scores. With this pipeline, we can test whether specialized modules can better capture score variation rather than relying on a single model. Each component in this proposed pipeline is based on Gemma $4$ E$4$B, but LoRA fine-tuned for its specific task.

In order to understand the upper bound of our proposed router pipeline, we also evaluate with an oracle router. Here, each crossing is automatically routed to the correct regressor instead of using our classifier. This allows us to measure how well the specialist regressors could perform if the router decisions were correct, further helping us identify where bottlenecks may occur, and if they come from the regression modules or the router. 





\subsection{Description-based Analysis}
\label{eq:description_based_analysis}

\begin{figure*}[h]
    \centering
    \includegraphics[width=0.9\linewidth]{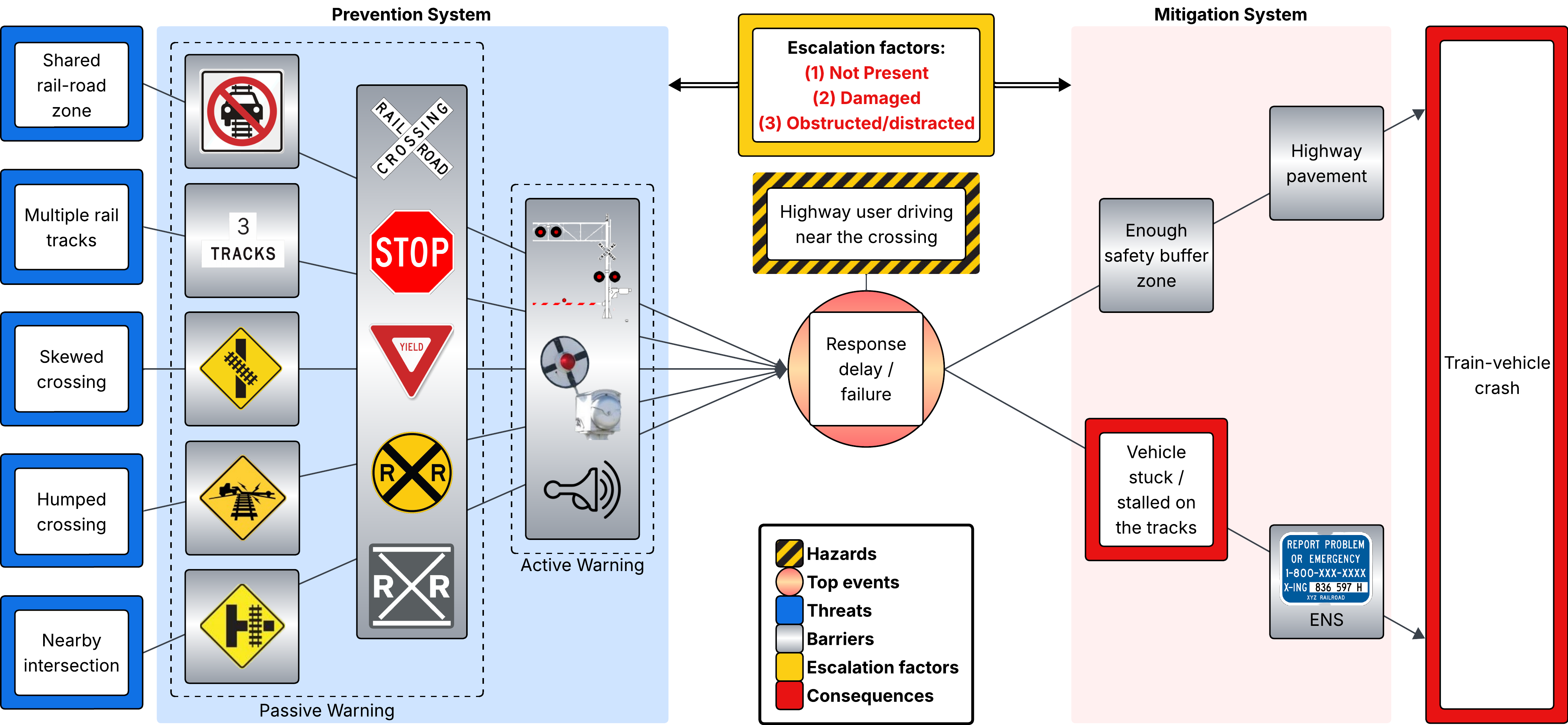}
    \caption{A bow-tie diagram that visualizes general sequences of crossing accidents. The left part indicates how barriers try to prevent threats from developing into response failure when hazard is present. The right part is how mitigation barriers work to keep from growing to worst consequences.}
    \label{fig:accident_theory}
\end{figure*}

\begin{figure}[ht]
    \centering
    \includegraphics[width=1\linewidth]{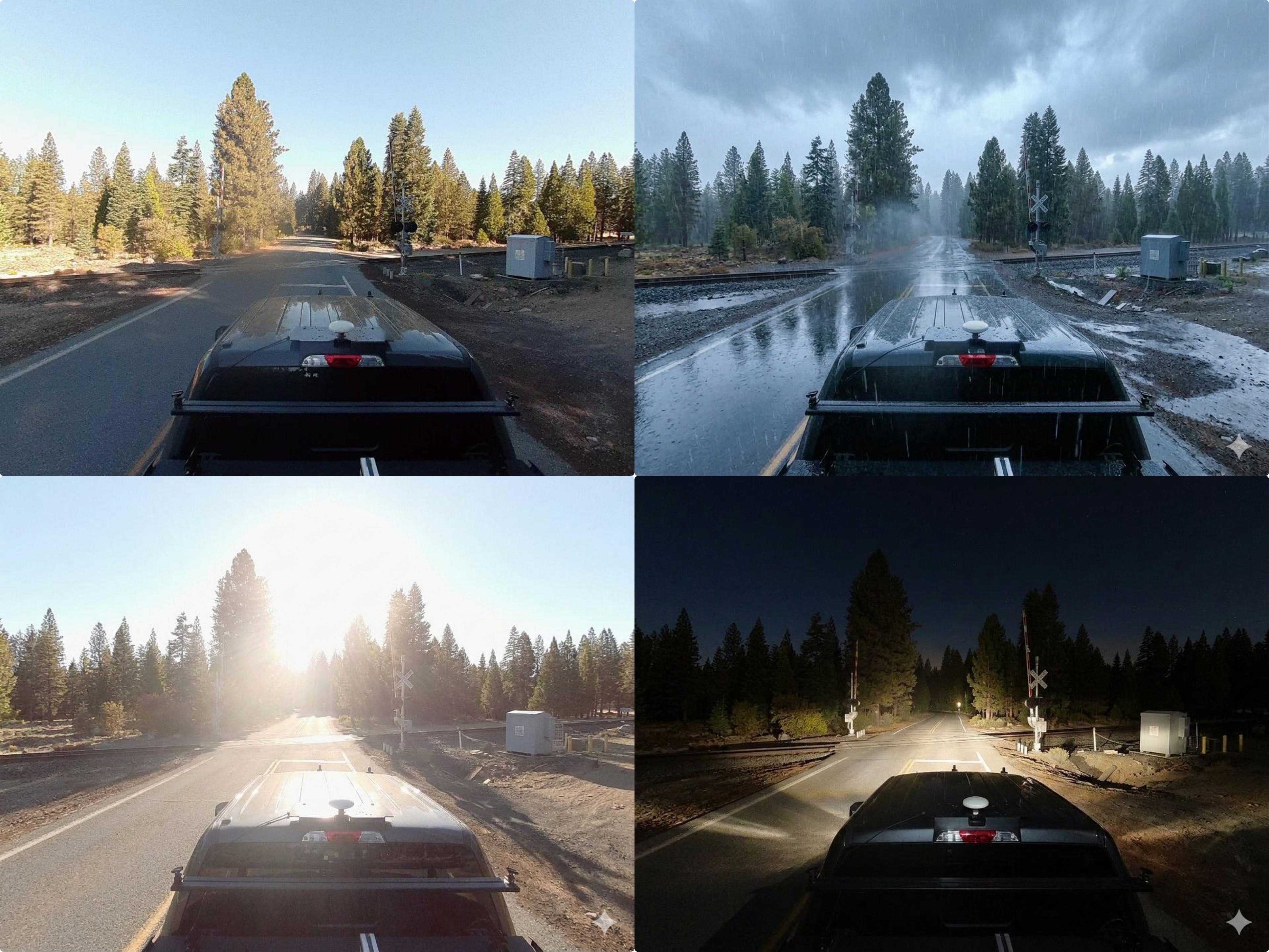}
    \caption{Sample augmented images. The top left is the original image, and the other three are augmented images: rainy weather, sunlight, and night.}
    \label{fig:augmentation}
\end{figure}

Description-based analysis aims to complement the score-based analysis by looking into other potential visual risk factors that are demonstrated only in images. This can uncover neglected aspects in the APS scoring function and provide more diversified insights to railway workers.

Instead of naively prompting VLMs to identify visual risks, we propose a systematic approach grounded on accident theory, specifically bowtie diagram, which is widely used in high-risk domains~\cite{bowtie_for_rail, bowtie_dynamic, bowtie_surgical}. As illustrated in Figure~\ref{fig:accident_theory}, the bowtie diagram centers around a \textit{top event}, which indicates a loss of control over a \textit{hazard} -- a source of potential harm, or "a highway user approaching near crossing" in our context. In association with the top event, the diagram consists of two main components: \textit{threats} and \textit{consequences}. Threats contribute to top event happening. For example, if the crossing is humped (threat), long trailers or limousines may be stuck on it (top event). When the control is lost, all energy of threats centers around the top event and results in certain consequences. In order to control unwanted scenarios and prevent threats from escalating into accidents, two types of \textit{barriers} are deployed: \textit{preventive} and \textit{mitigative}. However, there also exists what makes barriers fail -- \textit{escalation factors}, such as sunlight, shade, or vegetation that obscure or distract crossing warning signs from the driver's perspective. 

Our qualitative assessment is grounded in this framework. We guide a VLM inspect crossing images through the following steps:


\begin{enumerate}
    
    \item \textbf{Threat Identification}: Using our predefined list of threats (see Appendix~\ref{threat_list}), we find every structural and visual factors that correlate with threats from a sequence of images.
    \item \textbf{Barrier Identification}: Based on FRA glossaries and Form 57, we detect all safety infrastructures.
    \item \textbf{Escalation Factor Identification}: In the prevention system, we examine how environmental variances make drivers fail to recognize the safety infrastructures, based on our predefined risk (escalation) factors (see Appendix~\ref{escalation_list}). In the mitigation system which is related to how drivers would react to the pre-accident situation, we prompt the VLM to find out different types of response failure moments, such as poor highway pavement that makes breaking distance long.
    
\end{enumerate}

After inspecting each image, the VLM produces a structured output listing the crossing infrastructure visible in the scene and describing the status of each element.
\\


\noindent \textbf{Image Augmentation. } To account for the fact that crossing visibility and perceived risk can change under different environmental conditions, we evaluate each crossing under variations in illumination, shadows, weather, and time of day as shown in Figure~\ref{fig:augmentation}. To this end, we use \texttt{Gemini} to generate plausible variants of each image, such as rainy, sunny, and nighttime conditions. During generation, we enforce basic consistency constraints to preserve scene plausibility, such as ensuring that vehicle headlights are turned on at night.

\section{Results}
\label{sec:eval}

\subsection{Scoring}

\begin{table*}[t]
    \centering
    \caption{Binary risk classification results for E4B across held-out test splits. Results are reported as mean $\pm$ standard deviation across seeds.}
    \label{tab:binary-risk-classification}
    \begin{tabular*}{\textwidth}{@{\extracolsep{\fill}}lcccccc}
        \toprule
        \textbf{Model} 
            & \textbf{Accuracy} 
            & \textbf{Balanced Acc.} 
            & \textbf{High-Risk Prec.} 
            & \textbf{High-Risk Rec.} 
            & \textbf{High-Risk F1} 
            & \textbf{Macro F1} \\
        \midrule
        E4B No Report
            & $\underline{0.694 \pm 0.061}$
            & $0.491 \pm 0.068$
            & $0.180 \pm 0.106$
            & $0.175 \pm 0.087$
            & $0.176 \pm 0.096$
            & $0.494 \pm 0.067$ \\
        E4B Report
            & $0.623 \pm 0.073$
            & $0.559 \pm 0.055$
            & $0.235 \pm 0.052$
            & $0.460 \pm 0.077$
            & $0.308 \pm 0.056$
            & $\underline{0.523 \pm 0.055}$ \\
        E4B Oracle
            & $0.537 \pm 0.034$
            & $\underline{0.687 \pm 0.026}$
            & $\underline{0.270 \pm 0.012}$
            & $\mathbf{0.921 \pm 0.084}$
            & $\underline{0.417 \pm 0.018}$
            & $0.516 \pm 0.024$ \\
        E4B Fine-Tuned
            & $\mathbf{0.851 \pm 0.045}$
            & $\mathbf{0.767 \pm 0.063}$
            & $\mathbf{0.597 \pm 0.108}$
            & $\underline{0.635 \pm 0.124}$
            & $\mathbf{0.607 \pm 0.092}$
            & $\mathbf{0.757 \pm 0.059}$ \\
        \bottomrule
    \end{tabular*}
\end{table*}

\begin{table*}[t]
    \centering
    \caption{Continuous accident-score regression results for E4B across held-out test splits. Results are reported as mean $\pm$ standard deviation across seeds. The oracle routed regressor uses the ground-truth risk for routing and is included as an upper bound.}
    \label{tab:score-regression}
    \begin{tabular*}{\textwidth}{@{\extracolsep{\fill}}lcccc}
        \toprule
        \textbf{Model} 
            & \textbf{RMSE} 
            & \textbf{MAE} 
            & \textbf{Pearson Corr.} 
            & $\mathbf{R^2}$ \\
        \midrule
        E4B No Report
            & $0.562 \pm 0.195$
            & $0.309 \pm 0.069$
            & $-0.011 \pm 0.084$
            & $-95.915 \pm 91.674$ \\
        E4B Report
            & $0.432 \pm 0.069$
            & $0.315 \pm 0.046$
            & $0.014 \pm 0.149$
            & $-57.704 \pm 35.531$ \\
        E4B Oracle
            & $0.254 \pm 0.019$
            & $0.198 \pm 0.016$
            & $0.052 \pm 0.130$
            & $-18.545 \pm 11.818$ \\
        E4B Fine-Tuned
            & $\underline{0.071 \pm 0.029}$
            & $\underline{0.031 \pm 0.009}$
            & $0.245 \pm 0.215$
            & $-0.133 \pm 0.155$ \\
        Routed Regressor
            & $0.078 \pm 0.025$
            & $0.033 \pm 0.009$
            & $\underline{0.492 \pm 0.189}$
            & $\underline{0.067 \pm 0.375}$ \\
        Oracle Routed Regressor
            & $\mathbf{0.060 \pm 0.030}$
            & $\mathbf{0.019 \pm 0.007}$
            & $\mathbf{0.756 \pm 0.109}$
            & $\mathbf{0.536 \pm 0.181}$ \\
        \bottomrule
    \end{tabular*}
\end{table*}

We first evaluate whether E$4$B can identify high-risk crossings when framed as a binary classification task. As shown in Table~\ref{tab:binary-risk-classification}, when prompting the model with images alone, we get poor performance, specifically on the HIGH-RISK class. When we add the FRA Form $57$ records, we see an improvement of HIGH-RISK recall from $0.175$ to $0.460$, suggesting that the additional context helps the model more easily identify when crossing are HIGH-RISK. In the oracle prompting setting, where the model receives additional variables such as AADT, total trains, and train speed, we see the highest recall of $0.921$. This suggests that the model becomes much better at identifying HIGH-RISK crossings when it is given variables that are harder to infer. However, the model still has low precision on the HIGH-RISK class, which indicates that prompting alone may not be enough for this problem. Fine-tuning the E$4$B model achieves the best overall binary-classification results, achieving an accuracy of $0.851$, balanced accuracy of $0.767$, HIGH-RISK F$1$ score of $0.607$, and macro F$1$ of $0.757$.

Next, we evaluate the accident-score prediction task. As shown in Table~\ref{tab:score-regression}, we again see that the prompting-only baselines perform the worst at the task. Images alone, and even images with Form $57$ reports both produce large errors and extremely negative $R^2$ values. We see that the oracle-prompted model improves in RMSE, from $0.562$ for image-only prompting to $0.254$, but still has terrible correlation and $R^2$ value. This suggests that simply providing the model with the data, and information to calculate the accident-score is not enough. We need to find a way to allow the model to learn how to use this information to make much more calibrated numerical predictions.

This again leads us to fine-tuning. We see that the fine-tuned E$4$B model reduces RMSE to $0.071$ and MAE to $0.031$, which is much better than the prompting-only variants. However, its Pearson correlation is $0.245$ which is modest, and its mean $R^2$ is negative. This demonstrates the difficulty of the task, the target score distribution is very skewed, we observe that most of the crossing's accident scores are considered LOW-RISK crossings, with a smaller number being considered HIGH-RISK. This means that a model can achieve low error by simply predicting small values, while still struggling to properly calculate HIGH-RISK crossings. This suggests that we need a more robust method that can properly calculate both LOW-RISK, and HIGH-RISK crossings.

This finally leaves us with our routed regression pipeline. The learned routed regressor has a slightly worse RMSE and MAE than our single fine-tuned regressor, but it substantially improves our Pearson correlation from $0.245$ to $0.492$ and achieves a positive average $R^2$ of $0.067$. The oracle routed regressor further improves all metrics, achieving the best RMSE, MAE, Pearson correlation, and $R^2$. Since the oracle model does use the ground-truth risk group for routing, it should be interpreted as an upper-bound that we can hope to reach with a stronger router. The large gap between these routers indicate that the main bottleneck in the proposed pipeline is now the quality of the router rather than the regression models themselves.

Overall, these results show that fine-tuning is necessary for both binary risk prediction and score prediction. They also suggest that score prediction may benefit from a multi-module pipeline approach, rather than trying to be solved with one model.

\subsection{Qualitative Analysis}
\label{eq:qualitative_analysis}

\begin{figure}[ht]
    \centering
    \includegraphics[width=1\linewidth]{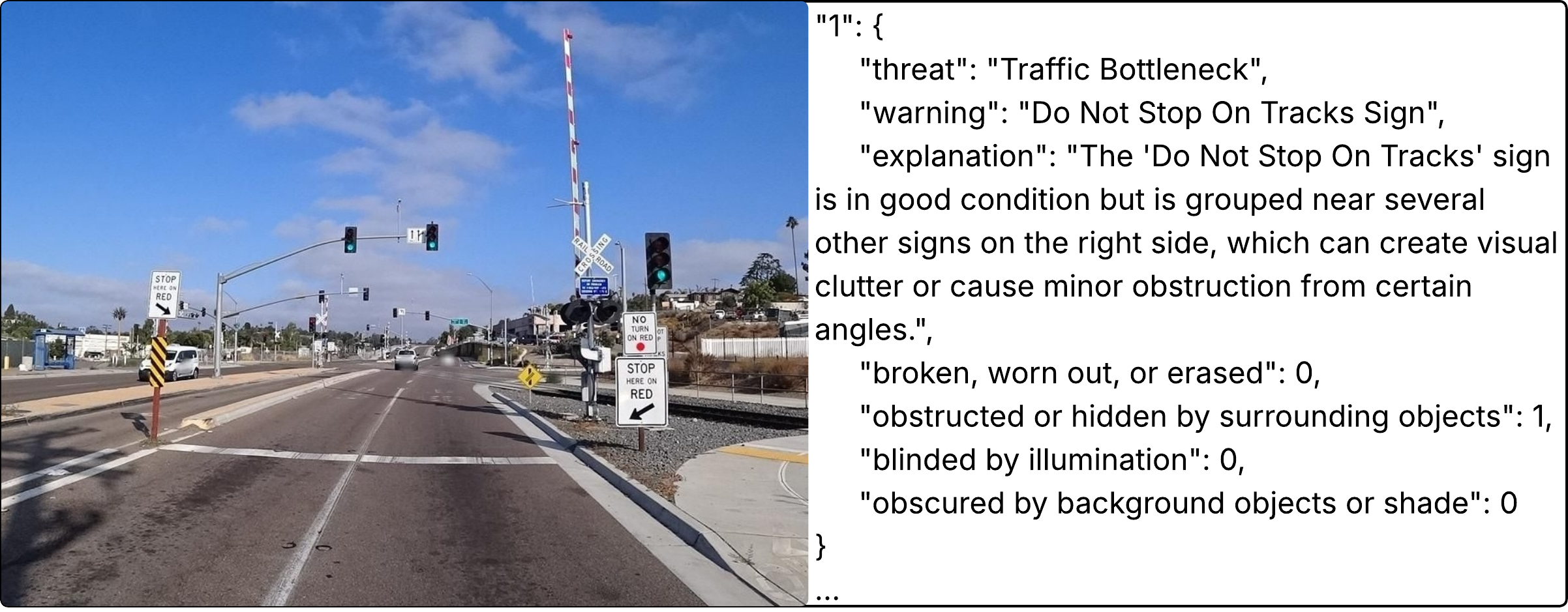}
    \caption{A sample input image and its output. Localized threats and crossing warning signs across a sequence of images were detected.}
    \label{fig:threat_barrier}
\end{figure}

\begin{table}
  \centering
  \caption{Detection success rates for threat and crossing warning signs over 10 executions}
  \label{tab:threat_barrier}
  \begin{tabular}{ll}
    \toprule
    Threats Present & Warnings Present \\
    \midrule
    Sharp-angled Crossing (1.00) & Skewed Crossing Sign (1.00) \\
    Multiple Lanes (1.00) & Cantilever FLS (0.00) \\
    \multirow{2}{*}{Traffic Bottleneck (1.00)} & Do Not Stop On Tracks Sign (0.80) \\
     & Keep Clear Pavement (1.00) \\
    \midrule
    - & FLS (1.00) \\
    - & Gate Arm (1.00) \\
    - & Crossbuck (1.00) \\
    \bottomrule
  \end{tabular}
\end{table}

Figure~\ref{fig:threat_barrier} illustrates a sample result where \texttt{Gemini-3.5-Flash} identified potential threats and their corresponding warning signs from the a sequence of images. In this scenario, the crossing lies at a sharp angle with the road, which may create blind spots for drivers; furthermore, multiple lanes might block warning sign visibility for drivers in the first lane, and traffic congestion may stall a vehicle on the track. As shown in Table~\ref{tab:threat_barrier}, the model successfully identified all the real threats perfectly, while presenting no hallucination regarding other threats. It also detected all predefined warning signs, missing the \textit{Do Not Stop On Tracks Sign} only 2 times out of 10 runs due to its being slightly obstructed by other signs from the driver's view.

\begin{table}
  \centering
  \caption{Comparison of escalation factor scores across augmented images. The highlighted values indicate risk factors that are expected to yield the highest scores.}
  \label{tab:escalation}
  \begin{tabular}{llrr}
    \toprule
    Warning Sign & Risk Factor & Image 1 & Image 2 \\
    \midrule
    \multirow{4}{*}{FLS} & broken & 0 & 0 \\
     & obstructed & 0.10 & 0.17\\
     & blinded & 0.10 & \textbf{2.83} \\
     & obscured & \textbf{1.00} & 1.50 \\
    \cmidrule{2-4}
    \multirow{4}{*}{Gate Arm} & broken & 0 & 0 \\
     & obstructed & 0.10 & 0.14 \\
     & blinded & 0 & \textbf{2.86} \\
     & obscured & \textbf{1.00} & 1.43 \\
    \cmidrule{2-4}
    \multirow{4}{*}{Crossbuck Sign} & broken & 0 & 0 \\
     & obstructed & 0 & 0.10 \\
     & blinded & 0 & \textbf{3.00} \\
     & obscured & \textbf{0.80} & 1.60 \\
    \bottomrule
  \end{tabular}
\end{table}

Table~\ref{tab:escalation} demonstrates how the VLM assesses the risk factors in the driver's perspective on the sample images (top left and bottom left) from Figure~\ref{fig:augmentation}. Apparently three warning signs are present at the crossing: flashing-light signals (FLS), gate arm, and crossbuck. We prompted the model to describe the extent to which each risk factor applies to the warning systems, scoring them on a 0–3 scale (0: not at all, 1: slightly, 2: moderately, 3: severely) based on how they affect a driver's ability to notice the crossing infrastructure. Given \textbf{Image 1} (top-left), the VLM judges that there is slight degradation in driver visibility, primarily due to warning signs being camouflaged by background objects, which aligns with human perception. However, in the \textbf{Image 2} (bottom-left), the VLM finds the blindness by the sunlight the most critical risk factor as it scores significantly high compared to other risk factors. Through these qualitative analyses, we illuminate the possibility of using VLMs for assessing risk factors in various scenarios.







\section{Conclusion}
\label{sec:conclusion}
In this work, we developed a novel multimodal pipeline for railway crossing risk assessment that combines street-level images, FRA accident records, and official APS scoring system. By leveraging reasoning capabilities of VLMs like \texttt{Gemma 4} or \texttt{Gemini}, we provide an alternative that can support scalable and automated expert inspections. Our quantitative evaluation demonstrates that addressing long-tailed data via router classifier and fine-tuning \texttt{Gemma 4} via LoRA significantly improve accident score prediction. Our framework also presents a possibility of identifying hidden visual risk factors across various augmented environmental conditions that were hard to be considered in the traditional safety analysis. While future work will expand this framework toward dynamic scenarios, such as video analysis and 3D simulation-based risk assessment, this proof-of-concept pipeline delivers a powerful approach to highway-rail grade crossing safety assessment for railway agencies. 

\section{GenAI Usage Disclosure}
Generative AI tools were used to assist with language polishing, grammar correction, and clarity improvements during manuscript preparation. Vision-language models are part of the methodology evaluated in this work, and the specific models used are described in the experimental setup. All AI-assisted writing was reviewed and edited by the authors, who take full responsibility for the final content of the manuscript.

\begin{acks}
    Research was supported by the University Transportation Center for Railway Safety (UTCRS) at UTRGV through the USDOT UTC Program under Grant No. 69A3552348340 and by the National Science Foundation under  grant no. 2431569, grant no. No. 2524228 and CREST Center for Multidisciplinary Research Excellence in CyberPhysical Infrastructure Systems (MECIS) grant no. 2112650.
\end{acks}

\appendix

\section{Pre-definition}

\subsection{Threat List}
As discussed in Section~\ref{eq:description_based_analysis}, our description-based analysis uses a predefined set of visual threats to guide the VLM inspection of railway crossing images. We specifically consider the following predefined threat categories:
\label{threat_list}
\begin{itemize}
    \item Sharp-angled crossing (creating driver blind spot)
    \item Humped crossing (causing low-clearance vehicles stuck)
    \item T-intersection before crossing (reducing reaction time)
    \item Multiple tracks (leading to secondary-train hit)
    \item Multiple lanes (making drivers miss warnings)
    \item Traffic bottleneck (stalling drivers on the track)
    \item Confusing rail-road zone (causing drivers enter the track)
    \item Parked vehicles beside the crossing (obstructing visibility)
\end{itemize}

\subsection{Escalation Factor List}
\label{escalation_list}
\begin{itemize}
    \item Broken, worn out, or erased
    \item Obstructed or hidden by surrounding objects
    \item Blinded by illumination
    \item Obscured by background objects or shade
\end{itemize}

\bibliographystyle{acm}
\bibliography{references}

\end{document}